\documentclass{article}

\usepackage{PRIMEarxiv}

\usepackage[utf8]{inputenc} % allow utf-8 input
\usepackage[T1]{fontenc}    % use 8-bit T1 fonts
\usepackage{hyperref}       % hyperlinks
\usepackage{url}            % simple URL typesetting
\usepackage{booktabs}       % professional-quality tables
\usepackage{amsfonts}       % blackboard math symbols
\usepackage{nicefrac}       % compact symbols for 1/2, etc.
\usepackage[nopatch=footnote]{microtype}
\usepackage{lipsum}
\usepackage{fancyhdr}       % header
\usepackage{graphicx}       % graphics
\usepackage{xcolor}
\usepackage[normalem]{ulem}

\usepackage{color}
\usepackage{amsmath}
\usepackage{amssymb}
\usepackage{setspace}

\usepackage{cite}
\usepackage{lineno}

\usepackage{bm}
\usepackage{multirow}
\graphicspath{{media/}}     % organize your images and other figures under media/ folder

\title{PINND: Physics Informed Neural Networks for Solving the Drift-Diffusion Equations
}

\author{
  Luke Ragan \\
  Department of Electrical \& Computer Engineering, \\
  Boston University \\
  Boston, MA, USA\\
  \And
  Matteo G.C. Alasio \\
  Dipartimento di Elettronica e Telecomunicazioni, \\ Politecnico di Torino, \\
  Torino, TO, Italy\\
  \And
  Enrico Bellotti \\
  Department of Electrical \& Computer Engineering, \\
  Division of Materials Science \& Engineering,\\
  Boston University, Boston, MA, USA\\
  \And
  Luca Dal Negro\textsuperscript{*} \\
  Department of Electrical \& Computer Engineering, \\
  Division of Materials Science \& Engineering,\\
  Physics Department, and Photonics Center \\
  Boston University, Boston, MA, USA\\
  \textsuperscript{*}\texttt{dalnegro@bu.edu} \\
}

\begin{document}
\maketitle

\begin{abstract}
The drift-diffusion equations are the fundamental building block for the modeling, design, and optimization of modern semiconductor devices. Solutions of the drift-diffusion equations rarely give way to analytical forms, and exhibit strong nonlinearity, with the solutions ranging over several orders of magnitude. These attributes have led to research into numerical solvers, and while many commercial Technology Computer Aided Design (TCAD) packages now exist, they may require long simulation times, exhibit sensitivity to initial guesses, and rely on mesh generation. Further, these solvers are not well suited for inverse problems where the design parameters of a device are retrieved from its desired functional characteristics. To address all these challenges, this work demonstrates the usage of Physics Informed Neural Networks for the Drift-diffusion (PINND) equations with errors comparable to traditional simulation software using advanced training techniques to solve problems beyond the well-behaved equilibrium regime. For this study, we specifically choose an abrupt PN-junction in both one and two dimensions that features extreme gradients of the carrier concentrations at the Ohmic contacts and an abrupt discontinuity in the doping profile. Using the proposed PINN architecture, we are able to retrieve accurate  results for all observables under all bias conditions. Additionally, we apply the PINN solver towards the inverse problem of retrieving doping profiles directly from desired electric field distributions or electrostatic potentials under non-zero applied biases. Our work establishes a powerful PINNs-based approach for semiconductor device modeling with the ability to solve inverse problems under non-equilibrium conditions for
microelectronics research and discovery.
\end{abstract}

% keywords can be removed

\section{\label{sec:intro}Introduction}
The use of semiconductor materials is ubiquitous throughout modern life, where they are essential for integrated circuits, optoelectronic systems, power management technologies, and many other areas in the field of microelectronics \cite{sze_physics_2021}. Accurately modeling the transport of charge carriers in semiconductors on the microscale is essential for predicting device performance, energy efficiency, and noise parameters \cite{2016Ferry}. Such modeling is used in computing, communications, sensing technologies, and more. \cite{xie_designing_2024, zhao_review_2024, thennarasu_comprehensive_2025}

The semiconductor device industry only continues to scale and push towards more complex device geometries \cite{zhuang2026gallium}. As a result, modeling of semiconductors is essential, both for incorporating microscale effects and also to reduce prototyping and development costs. The governing set of equations for a large number of device geometries and scales are the drift-diffusion equations \cite{1985Markowich}. These equations are a set of coupled elliptic partial differential equations (PDEs) that describe the electrostatic potential and carrier transport throughout a device. The drift-diffusion model is utilized by the majority of commercial Technology Computer Aided Design (TCAD) software. The model consists of Poisson and continuity equations. Together with boundary conditions, they are able to determine a physical solution and predict both microscopic quantities, like the electrostatic potential and carrier densities, as well as macroscopic quantities, like the current at the terminals. Due to the inherent nonlinearity of these equations, analytical solutions are rarely possible and numerical methods are required \cite{1985Markowich, 2012Markowich, 1983BankRose}.

Several large TCAD software packages, such as Synopsys TCAD Sentaurus \cite{Sentaurus_Device_W-2024.09-SP1}, Silvaco Atlas \cite{silvaco_sftw}, and COMSOL Multiphysics \cite{comsol_sftw} rely on Finite Element Method (FEM) or Finite Volume Method (FVM) solvers. These tools are able to solve the drift-diffusion equations accurately for a large number of geometries and materials, represent the current state-of-the-art, and are used extensively in industry and academia for the design and optimization of semiconductor devices.

However, FEM and FVM methods are known to be dependent on initial guesses \cite{1983BankRose}, mesh generation \cite{wang2025finite}, and long run times. These problems are particularly clear when considering complex device geometries in higher dimensions. Importantly, these solvers are also not well-suited for inverse design problems, where device parameters are retrieved from desired performance characteristics, such as the spatial distribution of the electric field inside the device.

These challenges motivate the introduction of an alternative Physics Informed Neural Networks (PINNs) \cite{raissi_physics-informed_2019} computational approach that is capable of accurately solving both forward and inverse problems in the strongly coupled nonlinear problems that arise beyond the well-characterized equilibrium condition \cite{riganti_ddnet_2026}. 
The PINN methodology utilizes neural networks and automatic differentiation to solve complex partial differential equations that arise in many engineering contexts \cite{raissi_hidden_2020,chen_physics-informed_2020,chen_physics-informed_2021,chen_physics-informed_2022,riganti_auxiliary_2023,riganti_multiscale_2025,riganti2026multiscale}. By identifying the output of the neural network as the surrogate solution of the PDE, PINNs leverage automatic differentiation techniques to accurately calculate the residuals of the equations up to machine precision at collocation points distributed throughout the solution domain. The residuals of the PDE and its boundary conditions are incorporated into loss functions with suitable error norms and minimized using standard gradient-based optimization techniques \cite{raissi_physics-informed_2019}. By training the neural network to minimize these losses,  PINNs retrieve solutions that accurately satisfy the differential constraints of the physical problem while minimizing the error norms of the residuals, effectively solving the original PDE subject to the boundary conditions. This method has been demonstrated in a variety of different fields ranging from fluid mechanics to nanophotonics to phonon transport \cite{cai_physics-informed_fluid_2021, chen_physics-informed_2020, chen2025physics, chen_physics-informed_2022, riganti_auxiliary_2023, riganti2026multiscale}.

PINNs have several characteristics that set them apart from traditional TCAD solvers. To begin, PINN is a mesh-free method. While recent works have shown that PINN benefits from more regularly spaced samples, the solver is not restricted to keeping the same points throughout training or ensuring that a specific feature has a higher density of points. In fact, methods such as residual-based adaptive sampling (RAD) can be used to allow the solver itself to pick the points of the most interest instead of assigning them beforehand \cite{wu_comprehensive_2023}. Second, PINNs are not sensitive to parameter shifts, noisy data, or initial guesses. The networks in PINNs are randomly initialized and then the expressivity of the network is able to find the correct solution through optimization. Third, PINNs feature unique advantages when scaling to higher dimensions. Lastly, by training on available data, PINNs can minimize additional loss terms embedding unknown PDE parameters, thus efficiently solving inverse problems on the same footing of forward problems. Importantly, PINNs have been shown to be effective at solving inverse problems under noisy conditions \cite{karniadakis_physics-informed_2021, chen_physics-informed_2022, Ultrasound_2020}. 
For the drift-diffusion equations, these unique features of PINNs open the way to efficient design of complex device structures and time-dependent problems beyond the reach of traditional simulation methods.

There have been several attempts to apply PINNs towards semiconductor device modeling in previous works. Petrosyan and Yepiskoposyan created a hybrid technique to couple PINN with traditional solvers to drastically speed up the convergence time \cite{petrosyan_hybrid_2026}. Chankla et al. utilized PINNs with an additional training dataset for modeling SiC MOSFETs \cite{chankla_demonstration_2025}. Cai et al. produced a one-dimensional PINN solver for a pn-junction \cite{cai_multi-order_2024}, and Riganti et al. solved the drift-diffusion equations for an abrupt doping profile with extensions in 2D and inverse problems but was limited to equilibrium conditions \cite{riganti_ddnet_2026}.

In this paper, we demonstrate accurate solutions to both forward and inverse problems that arise far from the equilibrium condition and report marked advances in using PINNs for semiconductor device modeling. Through advances to the training process and PINNs architecture, we are able to solve the drift-diffusion equations for a pn-junction with an abrupt doping profile under any bias condition. Further, we show that our PINNs methodology can be extended to model two dimensional geometries and inverse problems.

This work is a significant step forward towards developing a general PINN-based solver that can be used for the automatic design of semiconductor devices and the discovery of new and more effective geometries in desired material platforms. Our solver overcomes limitations that plague both previous PINN frameworks and traditional solver methods.

\section{\label{sec:methods}Methods}
\subsection{The Poisson Drift-Diffusion Equations}
We consider the steady-state drift-diffusion equations
\begin{align}
    \varepsilon \nabla^2 \phi &= q (n-p-C), \label{eq:poisson} \\
    \nabla \cdot \mathbf{J_n} &= qR, \label{eq:n_cont}\\
    \nabla \cdot \mathbf{J_p} &= -qR, \label{eq:p_cont}\\
    \mathbf{J_n} &= q\mu_n(U_t\nabla n - n\nabla\phi), \label{eq:n_current}\\
    \mathbf{J_p} &= - q\mu_p(U_t\nabla p + p \nabla \phi) \label{eq:p_current}
\end{align}
Here Eq. \ref{eq:poisson} is the Poisson equation, Eqs. \ref{eq:n_cont} and \ref{eq:p_cont} are the continuity equations, and Eqs. \ref{eq:n_current} and \ref{eq:p_current} relate the current densities to the carrier concentrations and electrostatic potential. Here $n$ and $p$ are the electron and hole carrier densities respectively. $R$ represents the recombination term. For the purposes of our results, the Shockley-Read-Hall recombination term will be used, $\phi$ is the electrostatic potential, and $\mathbf{J_n}$ and $\mathbf{J_p}$ are the electron and hole current densities.

From \cite{1985Markowich}, the variables are rescaled by defining $s = \frac{x}{L}$, where L is the characteristic length of the device, $V_T = \frac{k_b T}{q}$ is the thermal voltage, and the normalized quasi-Fermi levels are $\psi = \frac{\phi}{V_T}, \quad \phi_n = \frac{E_{F_n}/q}{V_T}, \quad \phi_p = \frac{E_{F_p}/q}{V_T}$, we rewrite the equations for a one-dimensional (1D) case as follows:
\begin{align}
    \frac{1}{\Xi^2}\frac{d^2\psi}{ds^2} + (\hat{p} - \hat{n} + C) = 0, \\
    \frac{dJ_n}{ds} = \Omega_nR_n, \quad \frac{dJ_p}{ds} = -\Omega_pR_p \\
    J_n = -\hat{n}\frac{d \phi_n}{dx}, \quad J_p = -\hat{p}\frac{d\phi_p}{dx}
\end{align}
The normalized quantities used here are summarized in Table~\ref{tab:normalization}.
\begin{table}[ht]
    \centering
    \caption{Normalized quantities in the drift-diffusion equations.}
    \label{tab:normalization}
    \renewcommand{\arraystretch}{1.3}
    \begin{tabular}{ll}
        \toprule
        Quantity & Definition \\
        \midrule
        Thermal voltage & $V_T = k_B T / q$ \\
        Reference density & $N_{ref} = \max(N_A, N_D)$ \\
        Intrinsic density & $\hat{n}_i = n_i / N_{ref}$ \\
        Electron density & $\hat{n} = \hat{n}_i\, e^{\psi - \phi_n}$ \\
        Hole density & $\hat{p} = \hat{n}_i\, e^{\phi_p - \psi}$ \\
        Electron current density & $\mathbf{J_n} = -\hat{n}\, \nabla\phi_n$ \\
        Hole current density & $\mathbf{J_p} = -\hat{p}\, \nabla\phi_p$ \\
        Debye length & $L_D = \sqrt{\varepsilon V_T / (q N_{ref})}$ \\
        Poisson coefficient & $\Xi^2 = (L / L_D)^2$ \\
        Electron recombination coefficient & $\Omega_n = L^2 / (\mu_n V_T \tau_n)$ \\
        Hole recombination coefficient & $\Omega_p = L^2 / (\mu_p V_T \tau_p)$ \\
        Electron SRH rate & $R_n = \dfrac{\hat{n}\hat{p} - \hat{n}_i^2}{(\tau_p/\tau_n)(\hat{n} + \hat{n}_i) + (\hat{p} + \hat{n}_i)}$ \\
        Hole SRH rate & $R_p = \dfrac{\hat{n}\hat{p} - \hat{n}_i^2}{(\hat{n} + \hat{n}_i) + (\tau_n/\tau_p)(\hat{p} + \hat{n}_i)}$ \\
        \bottomrule
    \end{tabular}
\end{table}
The use of the variables $\psi$, $\phi_n$, and $\phi_p$ as the outputs of the neural network is referred to as the quasi-Fermi level formulation. 
In contrast with formulations with the log scaled carrier densities, the quasi-Fermi levels are bound within a smaller range and have a simpler functional shape, making the approximation through neural networks more accurate and efficient. However, while the quasi-Fermi level formulation offers significant advantages over log scaled carrier formulations, our network architecture additionally utilizes the Slotboom variables to improve the overall accuracy under forward bias. The Slotboom variables, which are designed to eliminate the nonlinearities and nearly diverging gradients in the quasi-Fermi levels, are defined from \cite{slotboom1973computer} as $u$ and $v$ in Eqn.~\ref{eq:slotboom}.
\begin{align}
u = e^{-\phi_n}, \quad v = e^{\phi_p} \label{eq:slotboom}
\end{align}
In order to scale the Slotboom variables in the optimal range for PINN, we normalize them normalized by their boundary values to ensure they always range between in the $(0,1)$ interval. The normalized variables $\tilde{u}$ and $\tilde{v}$ are specified in Eq. ~\ref{eq:slotboom_norm}
\begin{align}
    u &= 1 + (e^{V/V_T} - 1) \tilde{u}, \quad v = e^{-V/V_T} + (1-e^{-V/V_T})\tilde{v} \label{eq:slotboom_norm}
\end{align}
By utilizing the Slotboom variables, the charge carrier densities and current densities are able to be calculated with the following relations:
\begin{align}
    n = \hat{n}_i\, e^{\psi}u &\implies \mathbf{J_n} = - \hat{n}_i e^\psi (e^{V/V_T} - 1)\nabla\cdot \tilde{u} \\
    p = \hat{n}_i e^{-\psi} v &\implies \mathbf{j_p} = \hat{n}_i e^{-\psi} (1 - e^{-V/V_T}) \nabla\cdot \tilde{v}
\end{align}
The normalized Slotboom variables $\tilde{u}, \tilde{v} \in [0, 1]$ take the values $0$ and $1$ at the two Ohmic contacts for any bias, which removes the bias-dependent scale of $u$ and $v$ (which span the range $[1, E_V]$) from the network outputs. Because the currents are linear in $\nabla\tilde{u}$ and $\nabla\tilde{v}$, the continuity equations become linear second-order equations in the Slotboom variables for a given potential.

\subsection{Boundary Conditions}
Two types of boundaries appear in the devices considered here. At an Ohmic contact the semiconductor is assumed to be in local thermal equilibrium and electrically neutral. With $V_c$ the voltage applied to the contact, this fixes the potential and both quasi-Fermi levels,
\begin{align}
    \psi_L &= \frac{V}{V_T} - \frac{V_{bi}}{2}, \qquad \psi_R = \frac{V_{bi}}{2},   \notag \\
    \phi_{n,L} &= \phi_{p,L} = \frac{V}{V_T}, \qquad \phi_{n,R} = \phi_{p,R} = 0 \label{eq:ohmic}
\end{align}
where $L$ denotes the value at $s=0$ on the p-type region and $R$ denotes the value at $s=1$ on the n-type region. Here
\begin{equation}
    V_{bi} = \ln\!\left(\frac{N_A N_D}{n_i^2}\right)
    \label{eq:vbi}
\end{equation}
is the built-in potential in units of $V_T$ (for the symmetric junctions considered here, $N_A = N_D$). In terms of the normalized Slotboom variables, Eq.~\eqref{eq:ohmic} means $\tilde{u} = \tilde{v} = 0$ at one contact and $\tilde{u} = \tilde{v} = 1$ at the other. In higher dimensions, additional boundary conditions must be added on the other edges of the device. Insulating (Neumann) boundary conditions are imposed on every boundary that is not an Ohmic contact. These conditions force no current to leave the device (Eq.~\eqref{eq:neumann_current}) and the normal component of the electric field to vanish at the boundary (Eq.~\eqref{eq:neumann_psi}),
\begin{align}
    \hat{\mathbf{n}}\cdot\mathbf{J_n} = \hat{\mathbf{n}}\cdot\mathbf{J_p} = 0
    \;&\Longleftrightarrow\; \hat{\mathbf{n}}\cdot\nabla\phi_n = \hat{\mathbf{n}}\cdot\nabla\phi_p = 0, \label{eq:neumann_current} \\
    \hat{\mathbf{n}}\cdot\nabla\psi &= 0, \label{eq:neumann_psi}
\end{align}
where $\hat{\mathbf{n}}$ is the outward normal. In one dimension both ends of the device are Ohmic contacts and no Neumann (isolating) boundaries exist. The Ohmic conditions are imposed exactly by the construction of the network outputs (hard boundary conditions, Sec.~\ref{sec:hardbc}), so they require no loss terms. The Neumann conditions in two dimensions are imposed as an additional penalty term within the total loss function by evaluating the boundary condition at points along the geometrical edge of the device (Sec.~\ref{sec:training}).

\subsection{Network Architecture}
\begin{figure}[t]
    \centering
    \includegraphics[width=\textwidth]{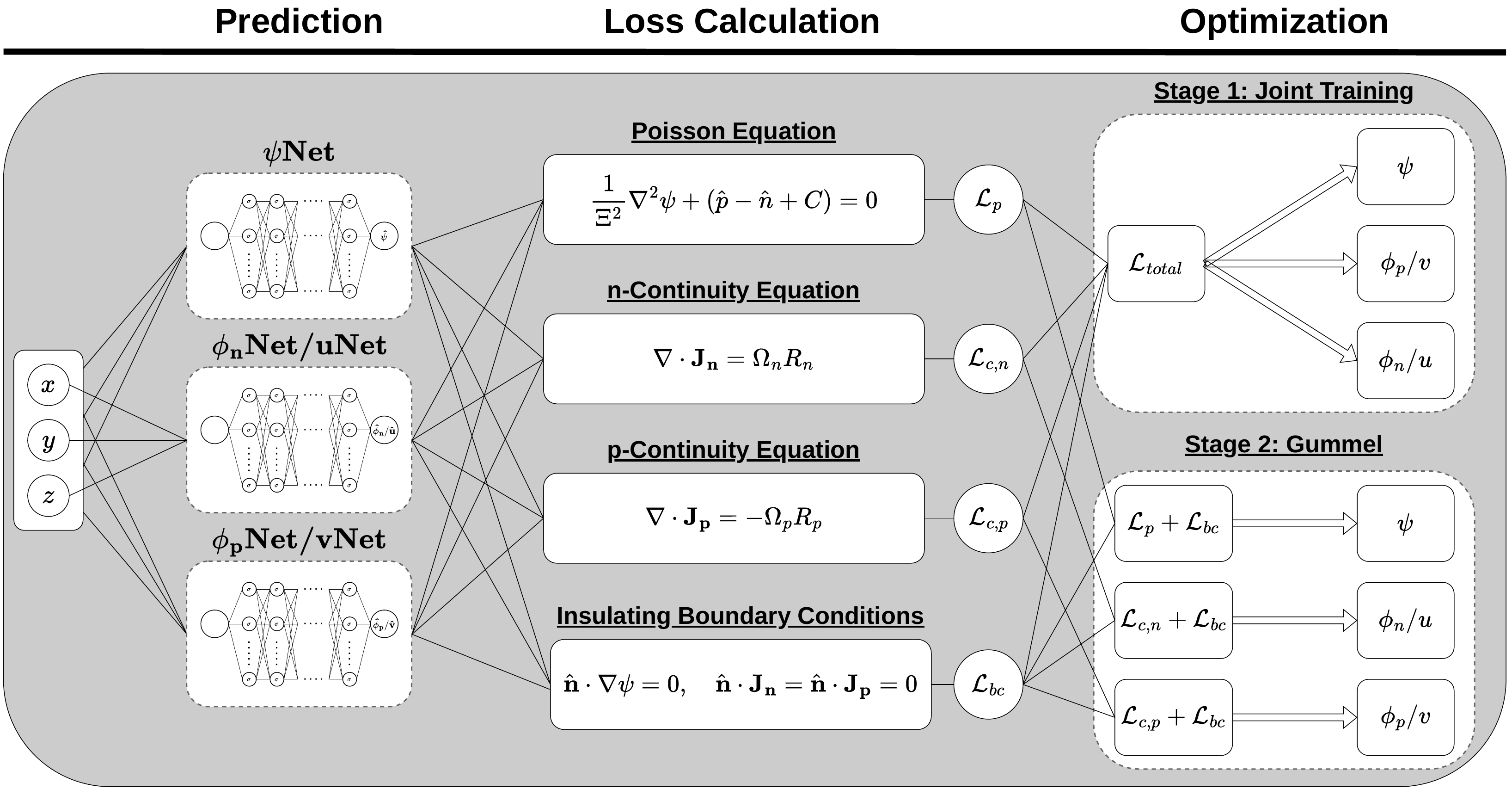}
    \caption{Structure of the PINN solver. Three fully connected networks predict the electrostatic potential $\psi$ and the two carrier variables (the quasi-Fermi levels $\phi_n$, $\phi_p$ or the normalized Slotboom variables $\tilde{u}$, $\tilde{v}$) from the position; the Ohmic contact values are built into the outputs exactly (Sec.~\ref{sec:hardbc}), so the loss contains no contact terms. Automatic differentiation of the outputs gives the residuals of the Poisson and continuity equations and, in two dimensions only, of the insulating (Neumann) boundary conditions, which form the loss terms $\mathcal{L}_p$, $\mathcal{L}_{c,n}$, $\mathcal{L}_{c,p}$ and $\mathcal{L}_{bc}$. In Stage 1 all networks minimize the total loss jointly; in Stage 2 (Gummel split) each network is updated only with the loss of its own equation and the boundary loss (Sec.~\ref{sec:gummel}).}
    \label{fig:network}
\end{figure}
Figure~\ref{fig:network} shows the structure of the developed PINN solver. Each unknown ($\psi$ and the two carrier variables, $\tilde{u}, \tilde{v}$ or $\phi_n, \phi_p$) is represented by its own fully connected network. Each network has 5 hidden layers of 32 units with sigmoid activations and a linear output, and takes the normalized position as input ($s$ in 1D problems, $x$ and $y$ in 2D). The networks are coupled only through the residuals of the governing equations. A systematic study of 2--6 hidden layers and 8--128 units per layer showed that the 5$\times$32 networks are the smallest architecture that is accurate for all observables at all biases tested; shallow networks (2--3 layers) frequently failed to converge. All networks are trained with the Adam optimizer \cite{kingma_adam_2017} ($\epsilon = 10^{-12}$) in double precision, with a learning rate that decays from $10^{-3}$ to $5\times10^{-5}$ on a cosine schedule over the full training. The baseline training uses 30\,000 epochs; the errors saturate after 20\,000--30\,000 epochs and any training beyond this point produced diminishing returns.

\subsubsection{\label{sec:hardbc}Hard boundary conditions}
In one dimension the Ohmic conditions are built into the outputs,
\begin{equation}
    f(s) = f_L (1-s) + f_R s + s(1-s)N_{\varphi}(s)
\end{equation}
where $f$ is a stand-in for each of the unknowns, $\mathcal{N}$ denotes the raw network output, and $\varphi_L$, $\varphi_R$ are the contact values of Eq.~\eqref{eq:ohmic}. By adding this extra step to the output of the network, the boundary values are then satisfied to machine precision for any network weights. Besides removing the boundary losses and their weights, this construction provides a physically reasonable starting point: the initial potential already spans the full range between the contacts, so the carrier densities, which depend exponentially on $\psi$, are of the correct order of magnitude from the first epoch.

In 2D the contacts are segments of the boundary, and the linear interpolation is replaced by a distance-function construction,
\begin{equation}
    f(x, y) = f_A\, w_A(x,y) + f_C\, [1 - w_A(x,y)] + D(x,y)\,\mathcal{N}_f(x,y), \qquad w_A = \frac{d_C}{d_A + d_C}, \quad D = \frac{d_A d_C}{d_A + d_C},
\end{equation}
where $f$ is any of the three unknowns, $f_A$ and $f_C$ its values at the Anode and Cathode, and $d_A$, $d_C$ smooth approximate distances to the two contact segments, constructed with R-functions \cite{shapiro2007semi}. The blending function $w_A$ is one on the Anode and zero on the Cathode, and $D$ vanishes on both contacts, so the contact values are again exact.

\subsection{\label{sec:training}Training Procedure}
The total loss is the sum of the mean square residuals of the governing equations at the collocation points $\{\mathbf{x}_i\}_{i=1}^{N_c}$ ($s$ is used in 1D, $x$ and $y$ in 2D) and of the insulating boundary conditions at the boundary points $\{\mathbf{x}_j\}_{j=1}^{N_b}$,
\begin{equation}
    \mathcal{L}_{total} = \mathcal{L}_p + \mathcal{L}_{c,n} + \mathcal{L}_{c,p} + \mathcal{L}_{bc},
    \label{eq:loss_total}
\end{equation}
with
\begin{align}
    \mathcal{L}_p &= \frac{1}{N_c}\sum_{i=1}^{N_c} \left[\frac{1}{\Xi^2}\nabla^2\psi + \hat{p} - \hat{n} + C\right]^2_{\mathbf{x}_i}, \qquad
    \mathcal{L}_{c,n} = \frac{1}{N_c}\sum_{i=1}^{N_c} r_n^2(\mathbf{x}_i), \qquad
    \mathcal{L}_{c,p} = \frac{1}{N_c}\sum_{i=1}^{N_c} r_p^2(\mathbf{x}_i), \\
    \mathcal{L}_{bc} &= \frac{1}{N_b}\sum_{j=1}^{N_b} \left[ \left(\hat{\mathbf{n}}\cdot\nabla\psi\right)^2 + \left(\hat{\mathbf{n}}\cdot\nabla\phi_n\right)^2 + \left(\hat{\mathbf{n}}\cdot\nabla\phi_p\right)^2 \right]_{\mathbf{x}_j},
\end{align}
Note that $\tilde{u}$, $\tilde{v}$ will take the place of $\phi_n$, $\phi_p$ in ~\eqref{eq:neumann_current} when the Slotboom formulation is used, and in 1D there are no insulating boundaries so $\mathcal{L}_{bc} = 0$. All derivatives are computed via automatic differentiation, and the resulting loss functions are scaled to ensure that none of the equations are neglected. In particular, two forms of the continuity residual are used. The unscaled form uses the continuity equation as written:
\begin{equation}
    r_n = \nabla\cdot\mathbf{J_n} - \Omega_n R_n, \qquad r_p = \nabla\cdot\mathbf{J_p} + \Omega_p R_p,
\end{equation}
Through the definition of $\mathbf{J_n}$ and $\mathbf{J_p}$, the residuals in this form are effectively weighted by the carrier densities at each location. This means that the residuals considering the current density when that carrier is the majority receive much higher attention. To address this issue, the prefactor form divides each continuity residual by the local carrier density, $\hat{n}$ for electrons and $\hat{p}$ for holes (in the Slotboom formulation, by the current prefactors $\hat{n}_i e^{\psi}(E_V - 1)$ and $\hat{n}_i e^{-\psi}(1 - E_V^{-1})$, which are proportional to $\hat{n}$ and $\hat{p}$ up to the bounded factors $u$ and $v$). The purpose of this operation is to enhance the residuals where electrons and holes are minority carriers. In the unscaled form, the residual of each continuity equation scales with the density of its own carrier, which in the minority region is smaller than in the majority region by up to $e^{V_{bi}}$, about 12 orders of magnitude in the 1D device. The minority-carrier residuals are then so tiny that they are effectively ignored by the loss, even though the minority-carrier currents determine the current through the junction. Dividing by $\hat{n}$ or $\hat{p}$ removes this disparity and weights all positions evenly, so that the minority-carrier regions contribute to the loss on an equal footing with the majority-carrier regions.

\subsubsection{\label{sec:gummel}Two-stage training with the Gummel split}
We consider two alternatives for training the networks. The first, which we use here for the forward-bias simulations, trains the networks with Adam in two stages. The motivation comes from the structure of the problem: the carrier densities depend exponentially on $\psi$, so the continuity equations are only meaningful once the potential, and with it the depletion region and the carrier landscape, is approximately correct. In the first, \emph{joint} stage (Stage~1 in Fig.~\ref{fig:network}) all three networks are trained on the total loss $\mathcal{L}_{total}$ of Eq.~\eqref{eq:loss_total}. In this stage the Poisson residual dominates, and the training predominantly establishes the electrostatics; the continuity residual is used in whichever of its two forms is smaller at initialization (in practice the unscaled form for $V \geq 0.15$\,V), so that it does not overwhelm the Poisson term. Once the potential is close to its final shape, continuing to train all networks on the summed loss becomes inefficient: the three residuals differ by orders of magnitude, and the gradient of the total loss with respect to each network is dominated by whichever equation is currently largest.

After 10\,000 epochs, training therefore switches to a \emph{Gummel split}, named after the decoupled iteration of Gummel \cite{gummel1964self} that is standard in TCAD solvers, where Poisson's equation is solved for the potential with the carrier densities held fixed and each continuity equation is then solved for its carrier with the potential held fixed. In the PINN analogue (Stage~2 in Fig.~\ref{fig:network}), each network is updated only by the gradient of its own equation and the boundary loss,
\begin{equation}
    \theta_\psi \leftarrow \nabla_{\theta_\psi}\left(\mathcal{L}_p + \mathcal{L}_{bc}\right), \qquad
    \theta_n \leftarrow \nabla_{\theta_n}\left(\mathcal{L}_{c,n} + \mathcal{L}_{bc}\right), \qquad
    \theta_p \leftarrow \nabla_{\theta_p}\left(\mathcal{L}_{c,p} + \mathcal{L}_{bc}\right),
\end{equation}
where $\theta_\psi$, $\theta_n$ and $\theta_p$ are the parameters of the $\psi$, electron and hole networks, with the continuity residuals now in the prefactor form, while all networks are still updated simultaneously at every step. Because each network sees only its own equation after the split, and Adam is invariant to a constant rescaling of the loss, the relative weighting of the loss terms no longer affects the updates; no loss weights need to be tuned. The split also prevents the electron and hole networks from being pulled by the Poisson residual, which they influence only through the carrier densities. We found this two-stage procedure essential: with joint training throughout, the continuity residuals stagnate, while after the split they decrease by several orders of magnitude.

\subsubsection{\label{sec:soap}Second-order preconditioning with SOAP}
The Adam optimizer scales the update of each parameter by a running estimate of the magnitude of its own gradient. This diagonal scaling treats every weight independently, while the curvature of the PINN loss couples the weights within each layer. When the loss is a sum of terms whose gradients point in competing directions, as is the case for the Poisson and continuity residuals, a diagonal preconditioner cannot resolve the conflict, and progress on one equation comes at the expense of the others~\cite{wang_gradient_2025}. With Adam, this is the reason the training is split into two stages (Sec.~\ref{sec:gummel}).

Another approach to addressing the conflicting gradients in the optimizer is to account for the correlations in the parameters when updating the weights. In its most expensive form, a full second-order preconditioner could be used to rescale the gradient with the inverse square root of the full matrix of gradient correlations. In our application (as well as many others), however, this simply requires too much storage to be practical. The authors of~\cite{gupta_shampoo_2018} introduced Shampoo for this very reason. As a play on the preconditioner name, Shampoo works by keeping one small preconditioner for each dimension of the weight matrix. With $G$ the gradient of a weight matrix $W \in \mathbb{R}^{m \times n}$, it accumulates the two factors
\begin{equation}
    L = \mathbb{E}\left[G G^T\right] \in \mathbb{R}^{m \times m}, \qquad R = \mathbb{E}\left[G^T G\right] \in \mathbb{R}^{n \times n},
\end{equation}
whose Kronecker product approximates the full correlation matrix, and updates the weights as
\begin{equation}
    W \leftarrow W - \eta\, L^{-1/4} G\, R^{-1/4}.
\end{equation}
Shampoo was shown to improve convergence~\cite{gupta_shampoo_2018}, but it is also very sensitive to hyperparameters and requires more calculation. As a more well-conditioned alternative, SOAP (ShampoO with Adam in the Preconditioner's eigenbasis)~\cite{vyas_soap_2024} was proposed to account for competing gradient directions while not significantly impacting convergence.

SOAP keeps $L$ and $R$ as exponential moving averages, together with their eigenvector matrices $Q_L$ and $Q_R$. The gradient is rotated into this basis, $G' = Q_L^T G\, Q_R$, the Adam moments are accumulated there, and the update is rotated back,
\begin{equation}
    \Delta W = -\eta\, Q_L \left( \frac{\hat{m}'}{\sqrt{\hat{v}'} + \epsilon} \right) Q_R^T,
\end{equation}
where $\eta$ is the learning rate and $\hat{m}'$ and $\hat{v}'$ are the bias-corrected first and second moments in the rotated basis. In this basis the curvature is close to axis-aligned, so the diagonal scaling of Adam approximates a full-matrix, Newton-like preconditioner. The bases are refreshed every 10 steps with one power iteration and a QR decomposition, and bias vectors are updated with standard Adam ($\beta_1 = 0.9$, $\beta_2 = 0.999$). Since the hidden layers are only $32 \times 32$, the additional cost of the eigendecompositions is negligible.

With SOAP, the three networks are trained jointly on the total loss for the entire training, without the Gummel split. The continuity and Poisson residuals are used in an \emph{equilibrated} form, in which the sum of the terms of each residual, $\sum_k t_k$, is divided by the gradient-stopped sum of their magnitudes,
\begin{equation}
    \tilde{r} = \frac{\sum_k t_k}{\mathrm{sg}\bigl[\textstyle\sum_k |t_k|\bigr] + \delta \bigl\langle \textstyle\sum_k |t_k| \bigr\rangle},
\end{equation}
where $\mathrm{sg}[\cdot]$ denotes a stopped gradient, $\langle \cdot \rangle$ the mean over the collocation points and $\delta = 10^{-3}$ a floor. Each residual is then bounded by one in magnitude, so that the three equations enter the joint loss on an equal footing.

In our testing, we saw the best results with the split training under forward bias and with the SOAP training under reverse bias. The forward- and reverse-bias results in Sec.~\ref{sec:results} are therefore obtained with these two procedures, respectively.

\begin{figure}[t]
    \centering
    \includegraphics[width=1.0\textwidth]{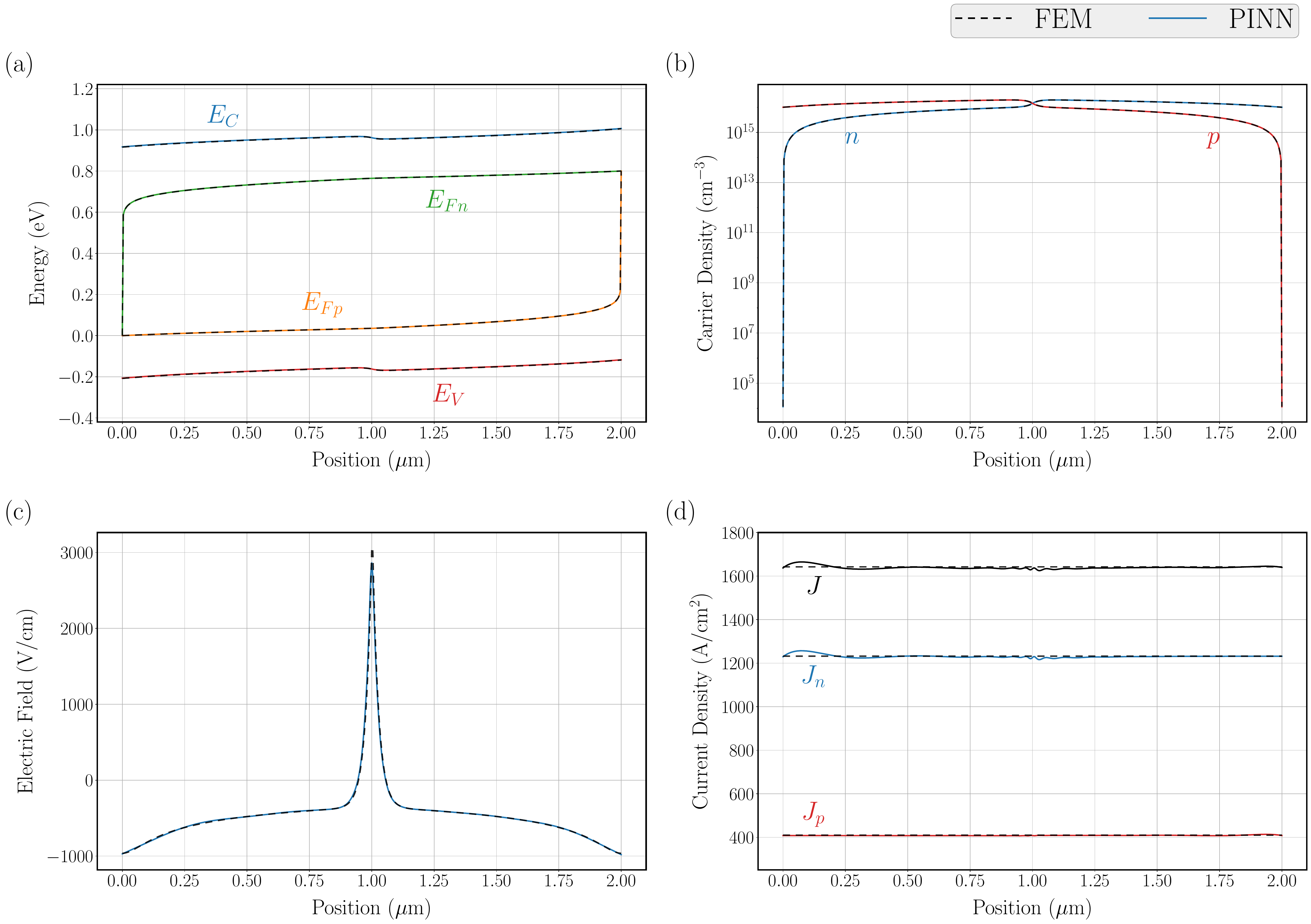}
    \caption{One-dimensional pn-junction ($L = 2\,\mu$m, $N_A = N_D = 10^{16}\,\mathrm{cm^{-3}}$) at a forward bias of 0.8\,V, solved with the normalized Slotboom variables and the two-stage training with the Gummel split (Sec.~\ref{sec:gummel}). (a) Band diagram with the conduction and valence band edges and the electron and hole quasi-Fermi levels, (b) electron and hole densities, (c) electric field and (d) electron, hole and total current densities. Solid lines: PINN; dashed lines: FEM. The relative $L_2$ errors, given in the boxes, are 0.26\% for $\psi$, 0.02\% and 0.01\% for $\log_{10} n$ and $\log_{10} p$, 0.03\% and 0.28\% for the quasi-Fermi levels and 0.45\% for the total current density. The steep minority-carrier boundary layers at both contacts (b) and the bend of the quasi-Fermi levels at the contacts (a) are resolved. The electric-field error (5.1\%) is concentrated at the junction peak, where the smooth doping step of the PINN differs from the abrupt profile of the reference.}
    \label{fig:forward08}
\end{figure}

\section{\label{sec:results}Results}
\subsection{One-dimensional pn-junction}
The first test case is an abrupt silicon pn-junction of length $L = 2\,\mu$m with $N_A = N_D = 10^{16}\,\mathrm{cm^{-3}}$ and the metallurgical junction at $x = 1\,\mu$m, at $T = 300$\,K ($n_i = 1.075\times10^{10}\,\mathrm{cm^{-3}}$, $\varepsilon_r = 11.7$, $\mu_n = 1417\,\mathrm{cm^2/Vs}$ and $\mu_p = 470.5\,\mathrm{cm^2/Vs}$, SRH lifetimes $\tau_n = 10\,\mu$s and $\tau_p = 3\,\mu$s). In the PINN solver the abrupt doping is represented by a smooth step,
\begin{equation}
    C(s) = N_D\,g(s) - N_A\,[1 - g(s)], \qquad g(s) = \tfrac{1}{2}\left[1 + \tanh\!\left(k\,(s - \tfrac{1}{2})\right)\right],
\end{equation}
with $k = 400$ (transition width of $\sim$5\,nm). The p-contact at $s = 0$ is grounded and the bias $V$ is applied to the n-contact at $s = 1$, so that Eq.~\eqref{eq:ohmic} gives the contact potentials $\psi_L = -V_{bi}/2$ and $\psi_R = V_{bi}/2 - V/V_T$, with $V_{bi} \approx 27.5$ (0.711\,V) from Eq.~\eqref{eq:vbi}. Reference solutions were computed with both an in-house Finite Element Method (FEM) solver and Sentaurus TCAD for identical parameters with an abrupt junction.
This problem is a demanding test for a PINN. Across the device the carrier densities vary by more than 12 orders of magnitude, and they do so in narrow regions: at the junction, the depletion region is a few tenths of a micrometer wide, and at the Ohmic contacts the minority densities collapse to their equilibrium values in boundary layers of a few nanometers. In addition, the doping changes sign over $\sim$5\,nm. A standard PINN such as that used in \cite{riganti_ddnet_2026} and \cite{cai_multi-order_2024}, with soft boundary conditions, the physical form of the residuals and a single summed loss, does not converge to the correct solution for this problem under forward bias: the loss is dominated by a few regions, and the steep gradients are either smeared or missed entirely.

We begin with the forward-bias regime, which has been the most difficult case for PINN solverse due to the extreme gradients at the boundaries. In particular, DDNet~\cite{riganti_ddnet_2026} was not able to solve the abrupt pn-junction under forward bias. Here, using the normalized Slotboom variables in conjunction with the two-stage training with the Gummel split (Sec.~\ref{sec:gummel}), the PINN converges to the correct solution: Figure~\ref{fig:forward08} shows the solution at a forward bias of 0.8\,V obtained with this procedure. The band diagram, the carrier densities, the electric field and the current densities all agree closely with the FEM reference. The relative $L_2$ errors with respect to the in-house FEM solver are 0.26\% for $\psi$, 0.02\% and 0.01\% for $\log_{10} n$ and $\log_{10} p$, 0.03\% and 0.28\% for the quasi-Fermi levels, and 0.45\% for the total current density. The steep boundary layers of the minority carriers at both contacts are resolved [Fig.~\ref{fig:forward08}(b)], as is the abrupt bend of the quasi-Fermi levels at the contacts [Fig.~\ref{fig:forward08}(a)]. The electric-field error (5.1\%) is concentrated at the junction peak, where the smooth doping step differs from the abrupt profile of the reference; with a softer step ($k = 80$) this error grows to 18\%, while the other observables are essentially unchanged.

\begin{figure}[t]
    \centering
    \includegraphics[width=\textwidth]{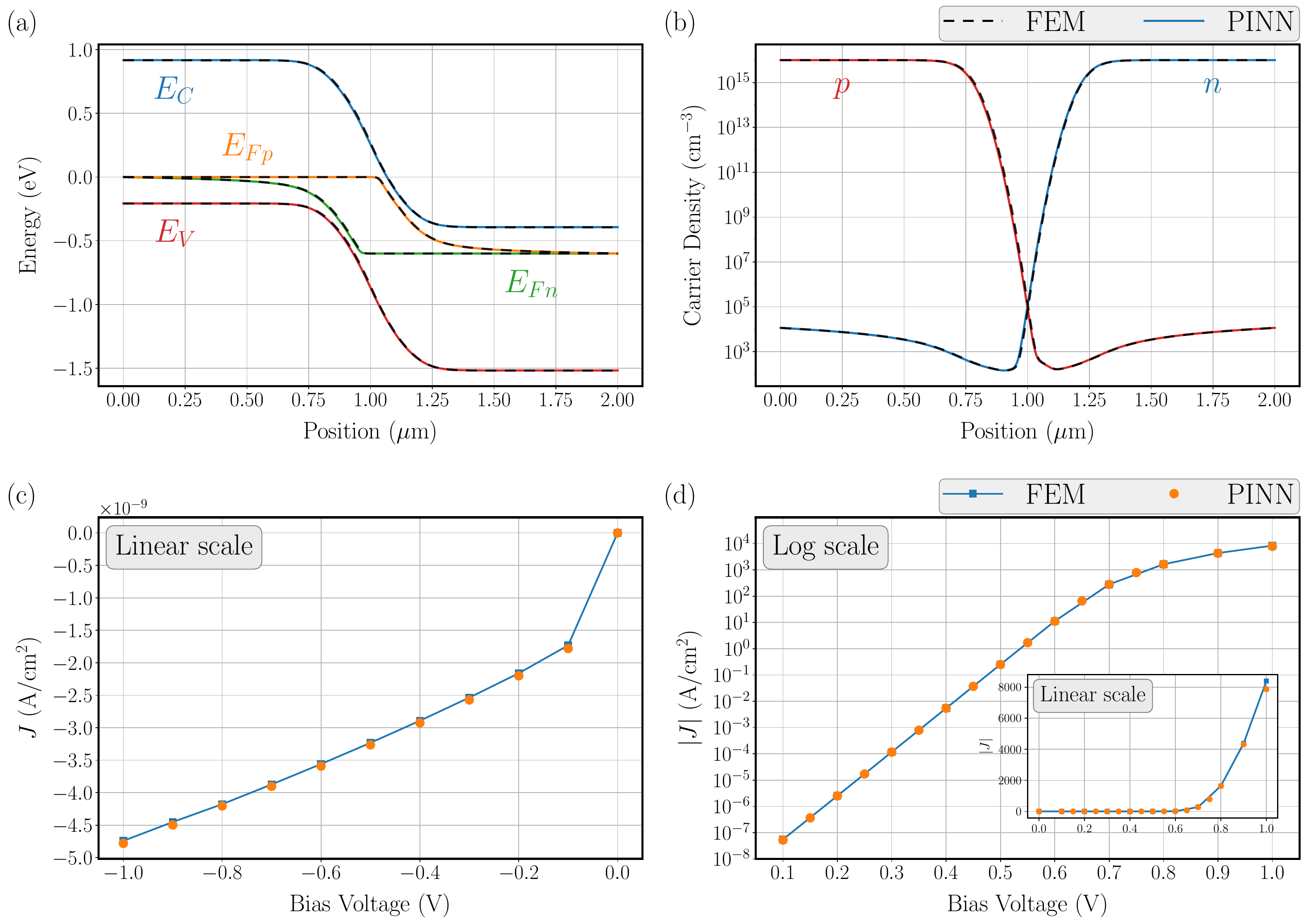}
    \caption{(a) Band diagram, with the conduction and valence band edges and the electron and hole quasi-Fermi levels, and (b) electron and hole densities of the one-dimensional pn-junction at a reverse bias of $-0.6$\,V, solved with the quasi-Fermi level variables and the SOAP training (Sec.~\ref{sec:soap}). Solid lines: PINN; dashed lines: FEM; the relative $L_2$ errors are given in the boxes. (c) Reverse-bias I-V characteristic between $-1.0$ and $-0.1$\,V on a linear scale and (d) forward-bias I-V characteristic between 0.1 and 1.0\,V on a logarithmic scale, with the same data on a linear scale in the inset; PINN terminal currents (circles) and FEM (squares and line), with the relative $L_2$ error of the terminal current over each curve given in the boxes. The reverse-bias currents are obtained with the SOAP training, where the runs at $-0.8$ to $-1.0$\,V are initialized from the solution at the neighbouring bias, and the forward-bias currents with the two-stage training (Sec.~\ref{sec:gummel}). The FEM currents at $-0.9$ and $-1.0$\,V are the median over the device, since the FEM solutions at these biases contain numerical current spikes.}
    \label{fig:reverse06}
\end{figure}

Under reverse bias, we demonstrate results at $-0.6$\,V in Fig.~\ref{fig:reverse06}(a) and (b). In this regime the depletion region widens, the minority-carrier densities near the junction fall far below their equilibrium values, and about two-thirds of the current at $-0.6$\,V is carried by thermal generation in the depletion region rather than by the diffusion of minority carriers. The solver must therefore resolve both the depletion region and the splitting of the quasi-Fermi levels across it. For these results we use the quasi-Fermi level formulation and the SOAP training method (Sec.~\ref{sec:soap}), with all three networks trained jointly on the equilibrated residuals for 30\,000 epochs. The relative $L_2$ errors with respect to FEM are 0.66\% for $\psi$, 2.4\% for the electric field, 0.50\% and 0.68\% for $\log_{10} n$ and $\log_{10} p$, 0.80\% and 0.50\% for the quasi-Fermi levels, and 0.73\% and 0.35\% for the electron and hole minority-carrier current densities, and the terminal current agrees with the FEM reference to within 0.9\%.

For the I-V characteristic, the terminal current must be extracted from the solution. Where a carrier is the majority carrier, its current density is the product of a large density and a very small quasi-Fermi level gradient, so the pointwise majority current is poorly conditioned. We therefore evaluate $J_n$ in the quasi-neutral p-region and $J_p$ in the quasi-neutral n-region, where each is the minority carrier, and account for the net recombination between the two regions,
\begin{equation}
    J = \langle J_n \rangle_p + \langle J_p \rangle_n + q \int_{s_p}^{s_n} R\, ds,
\end{equation}
which follows from integrating the continuity equations and holds for any doping profile. Figure~\ref{fig:reverse06}(d) shows the resulting forward-bias I-V characteristic between 0.1 and 1.0\,V. Between 0.2 and 0.8\,V the terminal current agrees with the FEM reference to within 1\%, over 9 orders of magnitude of current; the deviations are larger at 0.1\,V (5.0\%) and in high injection at 1.0\,V (6.3\%). The relative $L_2$ error over the whole forward-bias curve is 5.5\%, dominated by the largest currents at 0.9 and 1.0\,V. At reverse-bias [Fig.~\ref{fig:reverse06}(c)], the terminal current agrees with the FEM reference to within 1.5\% between $-0.3$ and $-1.0$\,V, and to within 2.8\% and 1.8\% at $-0.1$ and $-0.2$\,V. The relative $L_2$ error over the whole reverse-bias curve is 1.1\%.

Being able to solve the same device under both forward and reverse bias is an important step for PINN-based device simulation. The two regimes pose very different challenges. Under forward bias, carriers are injected across the junction and the minority-carrier densities in the neutral regions rise by many orders of magnitude, while under reverse bias the carriers are swept out of an ever wider depletion region and the current is set by thermal generation rather than by injection. The character of the solution therefore changes across the bias range, and with it the balance between the Poisson and continuity residuals in the loss. Importantly, a training procedure that converges in one regime does not necessarily converge in the other, which is why two training alternatives are used here. This strong dependence of the solution on the operating condition has been a major obstacle to applying PINNs to semiconductor devices, where the response to the applied bias is often the quantity of interest. Overcoming it opens the door to computing complete I-V characteristics with a single PINN solver, as shown below. Further, since the PINN is differentiable with respect to its inputs and parameters, it enables parametric and inverse studies across operating conditions, such as extracting device parameters from measured characteristics or designing doping profiles for a target response in both forward and reverse bias.

\subsection{Two-dimensional pn-junction}
Having demonstrated the accuracy of the PINN solver in the 1D case, we here extend the methodology to 2D device geometries. This is an important step since in realistic semiconductor devices the contacts often occupy only part of the surface, the junction has corners, and the current spreads laterally through the substrate. Therefore, a practical device solver must be able to handle multidimensional and complex geometries with mixed Ohmic and insulating boundaries. The mesh-free nature of PINNs makes this extension conceptually direct, since the networks simply take an additional input coordinate and the same losses are evaluated at collocation points distributed in the 2D plane.

\begin{figure}[t]
    \centering
    \includegraphics[width=\textwidth]{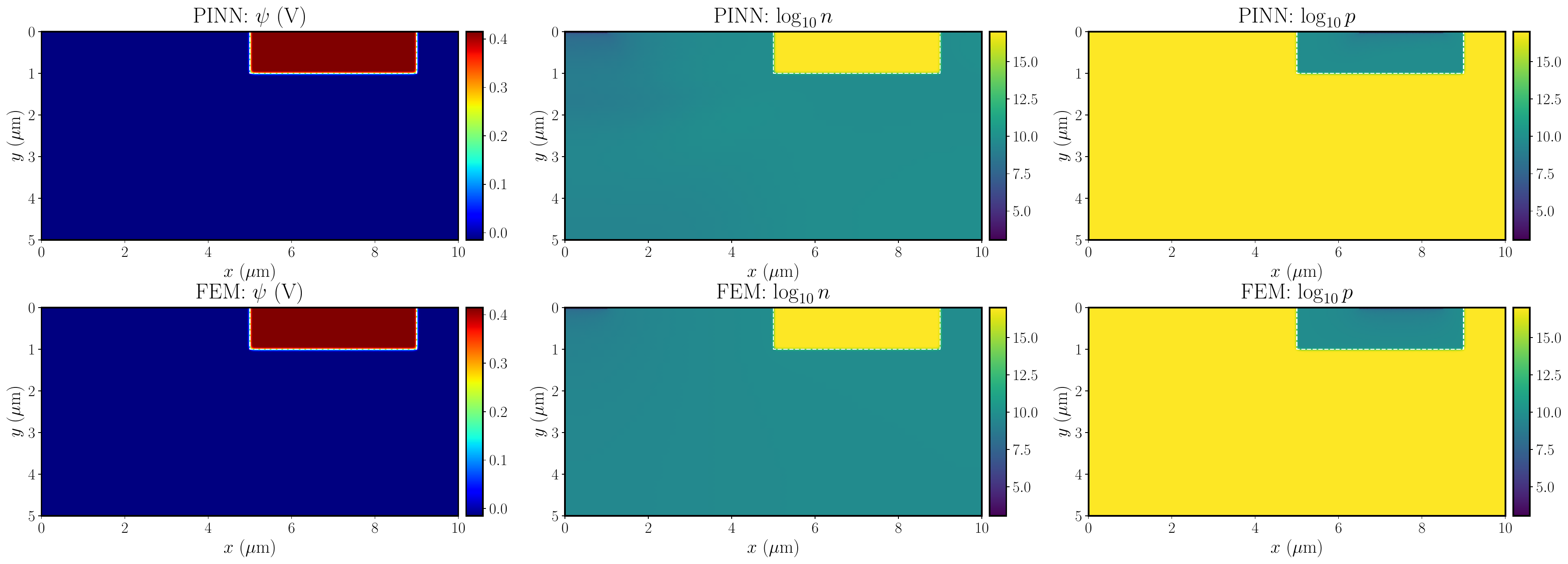}
    \caption{Two-dimensional pn-junction at 0.4V forward bias: a $10 \times 5\,\mu$m p-type substrate ($N_A = 10^{17}\,\mathrm{cm^{-3}}$) with an embedded $4 \times 1\,\mu$m n-type region ($N_D = 10^{17}\,\mathrm{cm^{-3}}$, dashed outline) at the top surface, contacted by the Anode ($0 \leq x \leq 1\,\mu$m) and the Cathode ($6.5 \leq x \leq 8.5\,\mu$m) with all other boundaries insulating. The PINN uses the quasi-Fermi level variables, the distance-function construction for the Ohmic contacts (Sec.~\ref{sec:hardbc}), the Neumann conditions as penalty terms, and collocation points taken from the FEM mesh. Columns: electrostatic potential $\psi$ (left), $\log_{10} n$ (center) and $\log_{10} p$ (right); rows: PINN (top) and FEM reference (bottom), on a common color scale per column. The relative $L_2$ errors are 2.08\% for $\psi$, 0.91\% for $\log_{10} n$ and 0.66\% for $\log_{10} p$; the depletion shell around the n-region and its corners are reproduced, and the remaining error is concentrated at the right edge of the n-region near the surface.}
    \label{fig:2d}
\end{figure}

The 2D silicon device considered here is a $10 \times 5\,\mu$m p-type substrate ($N_A = 10^{17}\,\mathrm{cm^{-3}}$) with an embedded n-type region ($N_D = 10^{17}\,\mathrm{cm^{-3}}$) of $4 \times 1\,\mu$m at the top surface, $5 \leq x \leq 9\,\mu$m (Fig.~\ref{fig:2d}). The Anode covers $0 \leq x \leq 1\,\mu$m on the p-type surface and the cathode $6.5 \leq x \leq 8.5\,\mu$m on the n-type surface and all other boundaries are insulating. Current therefore does not flow along a single axis. Under an applied bias it leaves the anode, crosses the substrate and enters the n-region through its bottom and side walls. The doping is a smooth version of the abrupt profile, $C(x, y) = (N_D + N_A)\, g(x - 5)\, g(9 - x)\, g(1 - y) - N_A$, with $g(z) = [1 + \tanh(z/\sigma)]/2$ and $\sigma = 5$\,nm. The reference solutions were computed using an in-house FEM solver and compared with Sentaurus TCAD.

Compared with the 1D case, the 2D problem introduces mixed boundary conditions. The Debye length is 13\,nm in a 10\,$\mu$m domain, the built-in potential is $\approx 32\,V_T$, and the depletion region is a thin shell of $\sim$0.1\,$\mu$m around the n-region, with field crowding at its two corners. The contact conditions are imposed with the distance-function construction of Sec.~\ref{sec:hardbc} and the Neumann conditions as additional penalty terms. At equilibrium the quasi-Fermi level variables are used, since the normalized Slotboom variables are undefined at $V = 0$. For this problem, we leverage the extensive FEM resources for mesh generation to choose our collocation points \cite{liu2014gmsh}. By carefully selecting points in this way, the PINN solver is able to converge faster and satisfy sharp features in the domain.

Figure~\ref{fig:2d} compares the PINN solution at 0.4V forward bias with the FEM reference. The potential and both carrier densities are reproduced across the device, including the depletion shell around the n-region and its corners, with relative $L_2$ errors of 2.08\% for $\psi$, 0.91\% for $\log_{10} n$ and 0.66\% for $\log_{10} p$. The exact imposition of the contact conditions proved decisive: with the same networks and soft contact losses, the error in $\psi$ was 18--41\%, because the initial potential is far from the contact values and the carrier densities start out at $n_i$ throughout the device, where the continuity equations are trivially satisfied.

These results demonstrate that our new training method can be naturally extended to higher dimensional problems, a variety of length scales, and more complex device geometries. The demonstrated ability to accurately model 2D geometries is a stepping stone towards applying the PINN methodology towards the complex heterostructures that currently utilized in the semiconductor industry.

\subsection{Inverse problems}
In many situations a designer may be interested in finding a doping profile that gives a desired electric field or electrostaic potential that is chosen purely based on design requirements. With current FEM methods, however, this can be a very time consuming and error-prone task. PINN, on the other hand, excels at solving inverse problems such as this one. Within the PINN framework, the unknown doping profile is simply added as an extra network output and a loss term is added on the given data to guide the optimization towards retrieving the correct doping profile, i.e., the one that minimizes, in a suitable norm, the error of the surrogate solution with respect to the training dataset. For the sake of demonstration, in what follows we consider two inverse problems for the 1D junction case: recovering the doping profile from the electric field $E(x)$, and from the electrostatic potential $\psi(x)$, each given under forward biased conditions.

In the introduced inverse problems, the doping profile is no longer a prescribed input but an additional output of the PINN. A fourth network $\mathcal{N}_C(s)$ (3 hidden layers of 32 tanh units, linear output) represents the doping $C(s)$, which replaces the prescribed profile in the Poisson residual. Since the Ohmic boundary conditions depend on the doping at the contacts, they are written in terms of this output.
\begin{equation}
    \psi(0) = \mathrm{asinh}\!\left(\frac{C(0)}{2\hat{n}_i}\right), \qquad \psi(1) = \mathrm{asinh}\!\left(\frac{C(1)}{2\hat{n}_i}\right) - \frac{V}{V_T},
    \label{eq:ohmic_inverse}
\end{equation}
This calculation reduces to Eq.~\eqref{eq:ohmic} for the true doping. The hard boundary construction of Sec.~\ref{sec:hardbc} uses these learned contact values, so that no information about the true doping enters the model. The data enter through an additional loss term, evaluated at $N_d = 2048$ points $\{s_k\}$ uniformly spaced in $s$ and interpolated from the FEM reference. For electric-field data and for potential data, respectively,
\begin{equation}
    \mathcal{L}_{data} = \frac{\frac{1}{N_d}\sum_{k} \left[\psi'(s_k) - \psi'_{data}(s_k)\right]^2}{\left\langle \psi'^{\,2}_{data} \right\rangle}, \qquad
    \mathcal{L}_{data} = \frac{\frac{1}{N_d}\sum_{k} \left[\psi(s_k) - \psi_{data}(s_k)\right]^2}{\mathrm{Var}\left(\psi_{data}\right)},
    \label{eq:loss_data}
\end{equation}
where $\psi' = d\psi/ds$ and $\psi'_{data} = -E_{data} L / V_T$ is the measured field in normalized units. The normalization by the mean square of the field, or by the variance of the potential, makes the data term of order one for either data type. The total loss of Eq.~\eqref{eq:loss_total} then becomes $\mathcal{L}_{total} = w_d\, \mathcal{L}_{data} + \mathcal{L}_p + \mathcal{L}_{c,n} + \mathcal{L}_{c,p}$, with a data weight $w_d = 10$.

The continuity residuals are used in the prefactor form, and the Poisson residual is pointwise equilibrated, i.e.\ divided by the gradient-stopped sum of the magnitudes of its terms, so that it does not reward an underestimated doping. Both inverse problems are solved with the normalized Slotboom variables and, importantly, outside of equilibrium: the data are taken under forward bias, at 0.5\,V for the electric field and at 0.4\,V for the potential. Training under bias is considerably harder than at equilibrium, since the doping must then be recovered together with the injected carrier densities and the current through the device, which couples the data, Poisson and continuity terms through the unknown doping.

\begin{figure}[ht]
    \centering
    \includegraphics[width=\textwidth]{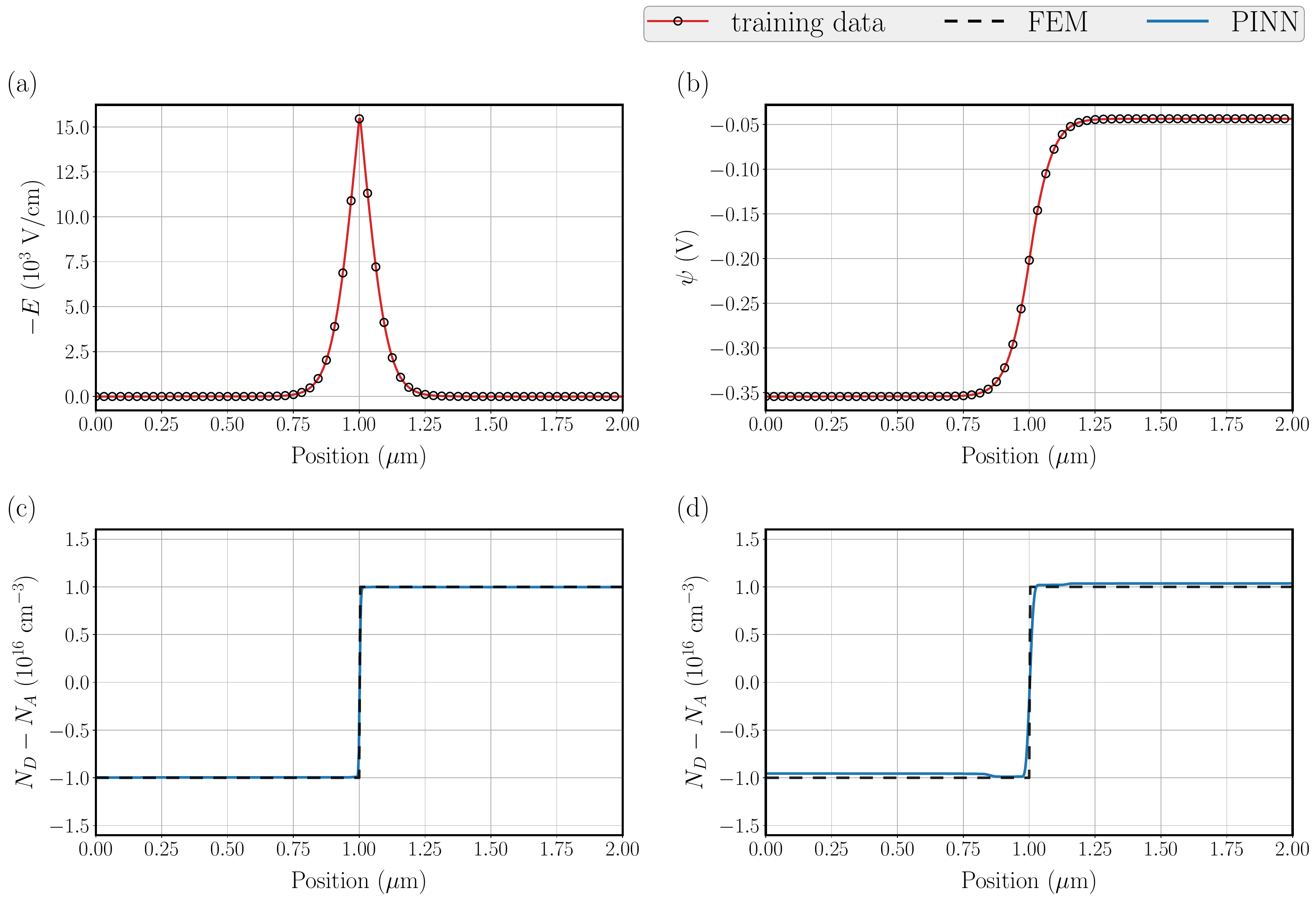}
    \caption{Doping profiles $N_D - N_A$ recovered by the inverse PINN (solid) compared with the true abrupt profile used in the FEM reference (dashed), (a) The electric field training data at a bias voltage of 0.5\,V. (b) The electrostatic potential training data at a bias voltage of 0.4\,V. (c) doping profile inversely retrieved from the electric field at 0.5\,V and (d) doping profile inversely retrieved from the electrostatic potential at 0.4\,V. The doping profiles have a relative $L_2$ errors of 3.89\% and 8.02\% for the electric field and electrostatic potential training respectively. For training on the electric field, Adam training refined with L-BFGS was used. Both plateaus are recovered to within 0.4\% and the junction to within 0.2\,nm. With the electrostatic potential, SOAP training followed by a continuation at a lower learning rate was used, and the junction was recovered to within 0.4\,nm, and the remaining error comes from an offset of the plateaus of about 4\%.}
    \label{fig:inverse}
\end{figure}

Figure~\ref{fig:inverse} shows the recovered doping profiles, together with the training data from which they are obtained. With Adam, the relative $L_2$ error of the doping is 5.4\% from the electric field at 0.5\,V and 9.6\% from the potential at 0.4\,V. Both reconstructions are further improved by refining the training. For the electric-field data, the networks are polished with L-BFGS~\cite{liu_limited_1989} on the same loss, with the loss weights held fixed, until convergence (about 13\,500 iterations), which reduces the error to 3.9\%. For the potential data, the networks are trained with SOAP (Sec.~\ref{sec:soap}) and then for a further 60\,000 epochs at a lower learning rate ($10^{-4}$ to $5\times10^{-6}$), which reduces the error to 8.0\%. From the electric field, the PINN recovers flat plateaus at $\mp10^{16}\,\mathrm{cm^{-3}}$ to within 0.4\% and a sharp step at the junction [Fig.~\ref{fig:inverse}(a)]. From the potential, the remaining error comes mostly from an offset of the plateaus of about 4\% [Fig.~\ref{fig:inverse}(b)]. In both cases the position of the junction is recovered to within 0.4\,nm.

\section{\label{sec:discussion}Conclusions}
In this work we introduced a novel and powerful PINN architecture that accurately solves the drift-diffusion equations in the context of both forward and inverse problems. This is a problem that has been notoriously difficult to tackle using traditional neural networks, as it involves an abrupt pn-junction under both forward and reverse bias, with carrier densities spanning more than 12 orders of magnitude and boundary layers only a few nanometer-thick at the contacts. Our results demonstrate that the main obstacle is not the expressivity of the neural networks but the conditioning of the optimization. The carrier densities and current prefactors depend exponentially on the potential, so residuals written in their physical form weight different regions of the device by many orders of magnitude, and the total loss of the coupled equations is dominated by whichever equation is currently largest. Exactly imposed Ohmic contacts, normalized residuals, and a training procedure that resolves the competition between the equations, either by decoupling them through a Gummel split or by preconditioning the joint optimization with SOAP, remove most of this disparity. Using small networks of only $\sim$5\,000 parameters each, our developed PINN the solver is capable of reproducing the FEM reference results with excellent accuracy, i.e., sub-percent errors in the potential, the carrier densities and the terminal current at both 0.8\,V and $-0.6$\,V, and follows the I-V characteristic over 9 orders of magnitude in forward bias and from $-0.1$ to $-1.0$\,V in reverse bias, in under 20 minutes of training on a single GPU.

The developed PINND framework was also extended to include 2D junction geometries with mixed contact and insulating boundaries. Finally, we demonstrated the solution of inverse problems for retrieving the doping profiles based on desired field distributions at the junctions. Since the unknowns of the problem enter only through the PDE and boundary residuals, the doping profiles can be recovered efficiently from the electric field or the electrostatic potential simply by making it a trainable function at the network output. Following this established procedure \cite{riganti_ddnet_2026}, we were able to achieve errors of 3.9\% and 8.0\% under non-equilibrium bias conditions, a task for which conventional PINNs and TCAD solvers are not designed.

This work establishes PINND as a powerful simulation tool in the field of semiconductor modeling and inverse design. Our advanced architecture and collocation strategy address several of the limitations of previous attempts and pave the way to numerous applications of PINNs for the solution of forward and inverse simulations in heterostructure devices under realistic operation conditions far from equilibrium, which are essential to next generation microelectronics research and automated discovery.

\section{Acknowledgments}
L. D. N. acknowledges the support from the U.S. Army Research Office, RF-Center managed by Dr. T. Oder (Grant \#W911NF-22-2-0158). The authors thank Anusha Krishnan and Roberto Riganti for insightful discussions. Numerical simulations were performed on the Shared Computing Cluster (SCC) at Boston University.

%Bibliography
\bibliographystyle{unsrt}  
\bibliography{references}  

@article{chen_physics-informed_2022,
	title = {Physics-informed neural networks for imaging and parameter retrieval of photonic nanostructures from near-field data},
	volume = {7},
	issn = {2378-0967},
	url = {https://doi.org/10.1063/5.0072969},
	doi = {10.1063/5.0072969},
	number = {1},
	urldate = {2023-05-26},
	journal = {APL Photonics},
	author = {Chen, Yuyao and Dal Negro, Luca},
	month = jan,
	year = {2022},
	pages = {010802},
}

@article{raissi_physics-informed_2019,
	title = {Physics-informed neural networks: {A} deep learning framework for solving forward and inverse problems involving nonlinear partial differential equations},
	volume = {378},
	issn = {0021-9991},
	shorttitle = {Physics-informed neural networks},
	url = {https://www.sciencedirect.com/science/article/pii/S0021999118307125},
	doi = {10.1016/j.jcp.2018.10.045},
	urldate = {2023-06-02},
	journal = {Journal of Computational Physics},
	author = {Raissi, M. and Perdikaris, P. and Karniadakis, G. E.},
	month = feb,
	year = {2019},
	pages = {686--707},
}

@article{chen_physics-informed_2020,
	title = {Physics-informed neural networks for inverse problems in nano-optics and metamaterials},
	volume = {28},
	copyright = {© 2020 Optical Society of America},
	issn = {1094-4087},
	url = {https://opg.optica.org/oe/abstract.cfm?uri=oe-28-8-11618},
	doi = {10.1364/OE.384875},
	number = {8},
	urldate = {2023-06-08},
	journal = {Optics Express},
	publisher = {Optica Publishing Group},
	author = {Chen, Yuyao and Lu, Lu and Karniadakis, George Em and Dal Negro, Luca},
	month = apr,
	year = {2020},
	pages = {11618--11633},
}

@article{riganti_auxiliary_2023,
	title = {Auxiliary physics-informed neural networks for forward, inverse, and coupled radiative transfer problems},
	volume = {123},
	issn = {0003-6951},
	url = {https://doi.org/10.1063/5.0167155},
	doi = {10.1063/5.0167155},
	number = {17},
	urldate = {2024-02-01},
	journal = {Applied Physics Letters},
	author = {Riganti, R. and Dal Negro, L.},
	month = oct,
	year = {2023},
	pages = {171104},
}

@article{riganti_multiscale_2025,
	title = {Multiscale {Physics}-{Informed} {Neural} {Networks} for the {Inverse} {Design} of {Hyperuniform} {Optical} {Materials}},
	volume = {13},
	copyright = {© 2025 Wiley-VCH GmbH},
	issn = {2195-1071},
	url = {https://onlinelibrary.wiley.com/doi/abs/10.1002/adom.202403304},
	doi = {10.1002/adom.202403304},
	language = {en},
	number = {16},
	urldate = {2025-07-14},
	journal = {Advanced Optical Materials},
	author = {Riganti, Roberto and Zhu, Yilin and Cai, Wei and Torquato, Salvatore and {Dal Negro}, Luca},
	year = {2025},
	note = {\_eprint: https://advanced.onlinelibrary.wiley.com/doi/pdf/10.1002/adom.202403304},
	pages = {2403304},
}

@misc{kingma_adam_2017,
	title = {Adam: {A} {Method} for {Stochastic} {Optimization}},
	shorttitle = {Adam},
	url = {http://arxiv.org/abs/1412.6980},
	doi = {10.48550/arXiv.1412.6980},
	urldate = {2025-09-07},
	publisher = {arXiv},
	author = {Kingma, Diederik P. and Ba, Jimmy},
	month = jan,
	year = {2017},
	note = {arXiv:1412.6980 [cs]},
}

@article{karniadakis_physics-informed_2021,
    title = {Physics-informed machine learning},
    volume = {3},
    copyright = {2021 Springer Nature Limited},
    issn = {2522-5820},
    url = {https://www.nature.com/articles/s42254-021-00314-5},
    doi = {10.1038/s42254-021-00314-5},
    number = {6},
    urldate = {2023-06-02},
    journal = {Nature Reviews Physics},
    author = {Karniadakis, George Em and Kevrekidis, Ioannis G. and Lu, Lu and Perdikaris, Paris and Wang, Sifan and Yang, Liu},
    month = jun,
    year = {2021},
    pages = {422--440},
}

@misc{riganti_ddnet_2026,
    title = {{DDNet}: {A} {Unified} {Physics}-{Informed} {Deep} {Learning} {Framework} for {Semiconductor} {Device} {Modeling}},
    shorttitle = {{DDNet}},
    url = {http://arxiv.org/abs/2509.08073},
    doi = {10.48550/arXiv.2509.08073},
    urldate = {2026-03-30},
    publisher = {arXiv},
    author = {Riganti, Roberto and Alasio, Matteo G. C. and Bellotti, Enrico and {Dal Negro}, Luca},
    month = jan,
    year = {2026},
    note = {arXiv:2509.08073 [physics]},
}

@article{Ultrasound_2020,
  title={Physics-informed neural network for ultrasound nondestructive quantification of surface breaking cracks},
  author={Shukla, Khemraj and Di Leoni, Patricio Clark and Blackshire, James and Sparkman, Daniel and Karniadakis, George Em},
  journal={Journal of Nondestructive Evaluation},
  volume={39},
  number={3},
  pages={61},
  year={2020},
  publisher={Springer}
}

@article{raissi_hidden_2020,
	title = {Hidden fluid mechanics: {Learning} velocity and pressure fields from flow visualizations},
	volume = {367},
	shorttitle = {Hidden fluid mechanics},
	url = {https://www.science.org/doi/full/10.1126/science.aaw4741},
	doi = {10.1126/science.aaw4741},
	number = {6481},
	journal = {Science},
	author = {Raissi, Maziar and Yazdani, Alireza and Karniadakis, George Em},
	month = feb,
	year = {2020},
	note = {Publisher: American Association for the Advancement of Science},
	pages = {1026--1030},
}

@article{zhao_review_2024,
	title = {A {Review} of {Diamond} {Materials} and {Applications} in {Power} {Semiconductor} {Devices}},
	volume = {17},
	issn = {1996-1944},
	url = {https://www.ncbi.nlm.nih.gov/pmc/articles/PMC11278176/},
	doi = {10.3390/ma17143437},
	number = {14},
	urldate = {2025-08-26},
	journal = {Materials},
	author = {Zhao, Feiyang and He, Yongjie and Huang, Bin and Zhang, Tianyi and Zhu, Hao},
	month = jul,
	year = {2024},
	pmid = {39063728},
	pmcid = {PMC11278176},
	pages = {3437},
}

@article{thennarasu_comprehensive_2025,
	title = {A comprehensive review on the application of semiconductor nanometal oxides photocatalyst for the treatment of wastewater},
	volume = {27},
	issn = {1618-9558},
	url = {https://doi.org/10.1007/s10098-024-02960-6},
	doi = {10.1007/s10098-024-02960-6},
	number = {2},
	urldate = {2025-08-26},
	journal = {Clean Technologies and Environmental Policy},
	author = {Thennarasu, G. and Rajendran, Swethashree and Kalairaj, Ashmitha and Rathore, Hanumanth Singh and Panda, Rames C. and Senthilvelan, T.},
	month = feb,
	year = {2025},
	pages = {495--516},
}

@article{xie_designing_2024,
	title = {Designing semiconductor materials and devices in the post-{Moore} era by tackling computational challenges with data-driven strategies},
	volume = {4},
	copyright = {2024 Springer Nature America, Inc.},
	issn = {2662-8457},
	url = {https://www.nature.com/articles/s43588-024-00632-5},
	doi = {10.1038/s43588-024-00632-5},
	number = {5},
	journal = {Nature Computational Science},
	author = {Xie, Jiahao and Zhou, Yansong and Faizan, Muhammad and Li, Zewei and Li, Tianshu and Fu, Yuhao and Wang, Xinjiang and Zhang, Lijun},
	month = may,
	year = {2024},
	note = {Publisher: Nature Publishing Group},
	pages = {322--333},
}

@misc{silvaco_sftw,
    author = {{SILVACO}},
	title = {{SILVACO} {TCAD®}},
	url = {https://silvaco.com/tcad/},
    year = {2025}
}

@misc{comsol_sftw,
    author = {{COMSOL AB}},
	title = {{COMSOL} {Multiphysics®} {v. 6.0,} {Stockholm,} {Sweden}},
	url = {https://www.comsol.com},
    year = {2024}
}

@inproceedings{cai_multi-order_2024,
	address = {New York, NY, USA},
	series = {{DAC} '24},
	title = {Multi-order {Differential} {Neural} {Network} for {TCAD} {Simulation} of the {Semiconductor} {Devices}},
	isbn = {979-8-4007-0601-1},
	url = {https://doi.org/10.1145/3649329.3656215},
	doi = {10.1145/3649329.3656215},
	booktitle = {Proceedings of the 61st {ACM}/{IEEE} {Design} {Automation} {Conference}},
	publisher = {Association for Computing Machinery},
	author = {Cai, Zifei and Huang, Anaoxue and Xiong, Yifeng and Mu, Dejiang and Miao, Xiangshui and Wang, Xingsheng},
	month = nov,
	year = {2024},
	pages = {1--6},
}

@book{sze_physics_2021,
	address = {Hoboken, NJ, USA},
	edition = {Fourth edition},
	title = {Physics of semiconductor devices},
	isbn = {978-1-119-61800-3 978-1-119-42913-5},
	publisher = {Wiley},
	author = {Sze, S. M. and Ng, Kwok Kwok and Li, Yiming},
	year = {2021},
}

@book{2016Ferry,
        address = {Boca Raton, FL, USA},
	title = {Semiconductor {Transport}},
	isbn = {978-1-315-26754-8},
	publisher = {CRC Press},
	author = {Ferry, David K.},
	year = {2016},
}

@book{2012Markowich,
    address = {New York City, USA},
  title={Semiconductor equations},
  author={Markowich, Peter A and Ringhofer, Christian A and Schmeiser, Christian},
  year={2012},
  publisher={Springer Science \& Business Media}
}

@book{1985Markowich,
    address = {New York City, USA},
  title={The stationary semiconductor device equations},
  author={Markowich, Peter A},
  year={1985},
  publisher={Springer Science \& Business Media}
}

@article{1983BankRose,
  title={Numerical methods for semiconductor device simulation},
  author={Bank, Randolph E and Rose, Donald J and Fichtner, Wolfgang},
  journal={SIAM Journal on Scientific and Statistical Computing},
  volume={4},
  number={3},
  pages={416--435},
  year={1983},
  publisher={SIAM}
}

@MANUAL{Sentaurus_Device_W-2024.09-SP1,
ORGANIZATION={Synopsys, Inc.},
TITLE="Sentaurus Device User Guide. Version W-2024.09-SP1",
ADDRESS={Mountain View, CA},
MONTH=SEP,
YEAR=2024}

@article{wu_comprehensive_2023,
    title = {A comprehensive study of non-adaptive and residual-based adaptive sampling for physics-informed neural networks},
    volume = {403},
    issn = {0045-7825},
    url = {https://www.sciencedirect.com/science/article/pii/S0045782522006260},
    doi = {10.1016/j.cma.2022.115671},
    urldate = {2026-04-10},
    journal = {Computer Methods in Applied Mechanics and Engineering},
    author = {Wu, Chenxi and Zhu, Min and Tan, Qinyang and Kartha, Yadhu and Lu, Lu},
    month = jan,
    year = {2023},
    pages = {115671},
}

@article{chen_physics-informed_2021,
    title = {Physics-informed learning of governing equations from scarce data},
    volume = {12},
    copyright = {2021 The Author(s)},
    issn = {2041-1723},
    url = {https://www.nature.com/articles/s41467-021-26434-1},
    doi = {10.1038/s41467-021-26434-1},
    language = {en},
    number = {1},
    urldate = {2026-04-10},
    journal = {Nature Communications},
    publisher = {Nature Publishing Group},
    author = {Chen, Zhao and Liu, Yang and Sun, Hao},
    month = oct,
    year = {2021},
    pages = {6136},
}

@article{petrosyan_hybrid_2026,
	title = {Hybrid {PINN}-{TCAD} framework for sub-percent p--n junction simulation with spillover error quantification},
	volume = {25},
	number = {1},
	doi = {10.1007/s10825-025-02455-7},
	journal = {Journal of Computational Electronics},
	author = {Petrosyan, Alfred V. and Yepiskoposyan, Armen S.},
	year = {2026},
	pages = {12},
}

@article{chankla_demonstration_2025,
	title = {Demonstration of accurate {ID}-{VG} characteristics modeling in {SiC} {MOSFETs} using separated artificial neural networks with small training dataset},
	volume = {15},
	number = {1},
	doi = {10.1038/s41598-025-03005-8},
	journal = {Scientific Reports},
	author = {Chankla, Manopat and Chen, Bang-Ren and Singh, Shivendra Kumar and Chauhan, Yogesh Singh and Lee, Wen-Jay and Chen, Nan-Yow and Kanjanachuchai, Songphol and Wu, Tian-Li},
	month = may,
	year = {2025},
	pages = {18941},
}

@article{cai_physics-informed_fluid_2021,
	title = {Physics-informed neural networks ({PINNs}) for fluid mechanics: a review},
	volume = {37},
	number = {12},
	doi = {10.1007/s10409-021-01148-1},
	journal = {Acta Mechanica Sinica},
	author = {Cai, Shengze and Mao, Zhiping and Wang, Zhicheng and Yin, Minglang and Karniadakis, George Em},
	year = {2021},
	pages = {1727--1738},
}

@misc{vyas_soap_2024,
	title = {{SOAP}: Improving and Stabilizing {Shampoo} using {Adam}},
	doi = {10.48550/arXiv.2409.11321},
	author = {Vyas, Nikhil and Morwani, Depen and Zhao, Rosie and Kwun, Mujin and Shapira, Itai and Brandfonbrener, David and Janson, Lucas and Kakade, Sham},
	month = sep,
	year = {2024},
	note = {arXiv:2409.11321},
}

@misc{gupta_shampoo_2018,
	title = {{Shampoo}: Preconditioned Stochastic Tensor Optimization},
	doi = {10.48550/arXiv.1802.09568},
	author = {Gupta, Vineet and Koren, Tomer and Singer, Yoram},
	month = feb,
	year = {2018},
	note = {arXiv:1802.09568},
}

@inproceedings{wang_gradient_2025,
	title = {Gradient Alignment in Physics-informed Neural Networks: A Second-Order Optimization Perspective},
	booktitle = {Advances in Neural Information Processing Systems},
	author = {Wang, Sifan and Bhartari, Ananyae Kumar and Li, Bowen and Perdikaris, Paris},
	year = {2025},
	note = {arXiv:2502.00604},
}

@article{liu_limited_1989,
	title = {On the limited memory {BFGS} method for large scale optimization},
	volume = {45},
	number = {1},
	doi = {10.1007/BF01589116},
	journal = {Mathematical Programming},
	author = {Liu, Dong C. and Nocedal, Jorge},
	year = {1989},
	pages = {503--528},
}

@article{zhuang2026gallium,
  title={Gallium-Based Semiconductor Integrated Circuits: Past, Present, and Future for Emerging Optoelectronic Devices},
  author={Zhuang, Jiachang and Zhu, Minmin and Zhang, Haizhong and Wang, Weipeng and Lu, Xiaoqiang},
  journal={Advanced Functional Materials},
  volume={36},
  number={32},
  pages={e27869},
  year={2026},
  publisher={Wiley Online Library}
}

@article{wang2025finite,
  title={Finite element approach based numerical framework for device simulator},
  author={Wang, Da-Wei and Zhang, Qing and Wan, Hang and Zhao, Wen-Sheng},
  journal={IEEE Transactions on Computer-Aided Design of Integrated Circuits and Systems},
  volume={44},
  number={8},
  pages={3197--3207},
  year={2025},
  publisher={IEEE}
}

@article{chen2025physics,
  title={Physics-informed neural networks for topological metamaterial design and mechanical applications},
  author={Chen, Kangkang and Dong, Xingjian and Gao, Penglin and Chen, Qian and Peng, Zhike and Meng, Guang},
  journal={International Journal of Mechanical Sciences},
  volume={301},
  pages={110489},
  year={2025},
  publisher={Elsevier}
}

@article{riganti2026multiscale,
  title={A multiscale auxiliary PINN framework for generalized phonon transport},
  author={Riganti, Roberto and Ragan, Luke and Dal Negro, Luca},
  journal={Materials Today Physics},
  pages={102159},
  year={2026},
  publisher={Elsevier}
}

@article{shapiro2007semi,
  title={Semi-analytic geometry with R-functions},
  author={Shapiro, Vadim},
  journal={ACTA numerica},
  volume={16},
  pages={239--303},
  year={2007},
  publisher={Cambridge University Press}
}

@article{gummel1964self,
  title={A self-consistent iterative scheme for one-dimensional steady state transistor calculations},
  author={Gummel, Hermann K},
  journal={IEEE Transactions on electron devices},
  volume={11},
  number={10},
  pages={455--465},
  year={1964},
  publisher={IEEE}
}

@techreport{liu2014gmsh,
  title={GMSH},
  author={Liu, Ning and Yu, Wenbin},
  institution = {Purdue University},
  year={2014}
}

@article{slotboom1973computer,
  title={Computer-aided two-dimensional analysis of bipolar transistors},
  author={Slotboom, Jan W},
  journal={IEEE Transactions on Electron Devices},
  volume={20},
  number={8},
  pages={669--679},
  year={1973},
  publisher={IEEE}
}

\end{document}